# A Lightweight Deep Learning Model for Automatic Modulation Classification using Dual Path Deep Residual Shrinkage Network

Prakash Suman and Yanzhen Qu

Colorado Technical University,

Colorado Springs, CO, USA

prakash.suman@student.ctuonline.edu; yqu@coloradotech.edu

*Abstract*— **Efficient spectrum utilization is critical to meeting the growing data demands of modern wireless communication networks. Automatic Modulation Classification (AMC) plays a key role in enhancing spectrum efficiency by accurately identifying modulation schemes in received signals—an essential capability for dynamic spectrum allocation and interference mitigation, particularly in cognitive radio (CR) systems. With the increasing deployment of smart edge devices, such as IoT nodes with limited computational and memory resources, there is a pressing need for lightweight AMC models that balance low complexity with high classification accuracy. This paper proposes a low-complexity, lightweight deep learning (DL) AMC model optimized for resource-constrained edge devices. We introduce a dual-path deep residual shrinkage network (DP-DRSN) with Garrote thresholding for effective signal denoising and design a compact hybrid CNN-LSTM architecture comprising only 27,000 training parameters. The proposed model achieves average classification accuracies of 61.20%, 63.78%, and 62.13% on the RML2016.10a, RML2016.10b, and RML2018.01a datasets, respectively demonstrating a strong balance between model efficiency and classification performance. These results underscore the model's potential for enabling accurate and efficient AMC on-edge devices with limited resources.**

## I. INTRODUCTION

Spectrum is a limited and valuable physical resource, and its efficient utilization is essential to support the growing data demands of wireless communication networks. In the dynamic landscape of modern communication systems, maximizing radio spectrum efficiency is paramount. Automatic Modulation Classification (AMC) is a key technology that significantly contributes to this goal. By enabling receivers to accurately identify the modulation scheme of incoming signals, AMC supports dynamic spectrum allocation and optimizes the use of available radio frequencies. This capability is particularly critical in cognitive radio (CR) systems, where secondary users must access the spectrum without interfering with primary users [1]. Accurate modulation recognition allows systems to adapt transmission parameters to current channel conditions, thereby enhancing data throughput within the allocated bandwidth.

The study of AMC is vital for industries such as telecommunications, satellite communication, IoT, smart device development, and military applications. AMC aids in signal intelligence, electronic warfare, and secure communication [2]. The telecommunications industry benefits from AMC's role in enhancing spectrum efficiency, mitigating interference, and managing dynamic spectrum to support data growth in congested wireless environments[3]. Internet of Things (IoT) and smart device developers can use lightweight AMC models for efficient data transmission and energy-efficient communication.

Technologies, such as IoT, medical sensors, and smart home systems, have advanced rapidly. A growing number of small devices need to communicate information wirelessly in a complex, noisy environment. To meet the demand for real-time performance and low complexity, DL models must be developed for faster and more efficient AMC solutions targeting edge devices like IoT [4].

Previous studies have predominantly focused on using more intricate neural network structures to extract diverse signal features, thereby improving recognition accuracy. However, this has resulted in larger model sizes and increased computational demands, requiring hundreds of thousands to millions of parameters, which are unsuitable for resource-constrained edge receivers. Simultaneously, some efforts to simplify DL-based models often compromise recognition accuracy, making it challenging to meet the requirements of edge devices in modern communication systems.

This paper proposes a low-complexity, lightweight DL AMC model for edge network devices with limited memory and computing resources. The key contributions of this work are as follows:

- We introduce the application of a dual-path deep residual shrinkage network (DP-DRSN) for AMC.
- We create an efficient hybrid convolution neural network (CNN) and Long Short-Term Memory (LSTM) model using only 27k training parameters.

- We enhance AMC recognition accuracy by integrating a self-learnable scaling approach for signal denoising thresholds, leveraging Garrote thresholding to balance denoising effectiveness, model performance, and computational complexity.

We trained and experimented with standard datasets, RML2016.10a, RML2016.10b [5], and RML2018.01a [6], which have been widely used in prior AMC research. The proposed model uses only 27k parameters, 2.3 million to 18.82 million floating-point operations per second (FLOPs), and is 108.83 kB in size, achieving 1.04ms inference time per sample and 31.5mJ of energy usage per sample. Using the RML2016.10a, RML2016.10b, and RML2018.10a datasets, it achieves an average classification accuracy of 61.20%, 63.78%, and 62.13%, respectively.

This paper is organized as follows: Section II discusses related work, Section III presents the problem statement, hypothesis, and research question, Section IV outlines the methodology and the proposed AMC model, Section V details the experiment, analysis of results, and key takeaways, Section VI provides discussion, and recommendations for future research, and Section VII concludes the paper.

## II. RELATED WORK

The need for lightweight AMC models to support edge devices, such as IoT and CR systems, is becoming increasingly critical. However, balancing computational efficiency and classification accuracy remains a significant challenge in resource-constrained, interference-prone, and spectrum-dense environments. Several recent studies have proposed innovative approaches to address these challenges.

### A. Models with <50k Tunable Parameters

Lin et al.[4] introduced a novel lightweight AMC model using Liquid State Machine (LSM) spiking neural networks (SNNs) with 7,120 parameters, achieving faster processing but lower average accuracy (36.69%-79.68%) for RML2016.10a, RML2016.10b, and RML2018.01a datasets. In [7], Shaik and Kirthiga used DenseNet with 19k parameters, showing 55%-70% accuracy for RML2018.01a. TianShu et al.'s [8] Long-term Features Neural Network (IQCLNet) classifier with I/Q correlator had 29k parameters, reaching 59.73% accuracy for RML2016.10a. Shen et al. [9] developed a multi-subsampling self-attention network (MSSA) using dilated convolution branches with self-attention, with variants from 36k to 218k parameters, achieving 55.25%-60.90% accuracy for Unmanned Aerial Vehicle (UAV)-to-ground communication systems. An et al. [10] proposed the Threshold Denoise Recurrent Neural Network (TDRNN) model. The TDRNN consists of two key components: a Threshold Denoiser (TD) module using soft thresholding to filter out noise from received signals and a GRU layer to classify the denoised signals. The TDRNN model has 41k parameters and an average accuracy of 63.5%, using the RML2016.10a dataset.

Gao et al. [11] built upon their previous work in [12] and developed a lightweight modulation recognition algorithm based on a CNN-LSTM dual-channel model, addressing class imbalance through data preprocessing. The algorithm reduced parameters from 44,379 to 43,941, achieving 66.7% accuracy for five selected modulations on the balanced RML2016.10a dataset and 64.2% in imbalanced scenarios (20:1 ratio). Li et al. [13] proposed a complex-valued transformer (CV-TRN) model using complex multi-head self-attention. With 44,790 parameters, CV-TRN achieved 63.74% accuracy on RML2016.10a and 64.13% on RML2018.01a. However, its reliance on relative position embedding and phase offset data augmentation adds complexity to its deployment. Su et al. [14] proposed SigFormer, a robust signal classification model using a pyramid Transformer architecture. SigFormer achieved 63.71% (RML2016.10a), 65.77% (RML2016.10b), and 63.96% (RML2018.01a) with models of 44k, 44k, and 158k parameters, respectively.

### B. Models with 50k to 100k Tunable Parameters

Ding et al. [15] addressed the limitations of pure data-driven AMC methods by incorporating expert knowledge to improve recognition in high-noise conditions. They used CNN, specifically ResNet, with a Bidirectional GRU (BiGRU), with dual-driven schemes combining data-driven and semantic knowledge approaches. With 69k parameters, the model achieved 100% accuracy at> 2dB SNR and 34% at -10dB for the RML2018.01a dataset. Zhang et al. [16] developed a model using CNN and GRU layers for feature extraction and achieved 60.44%-63.82% accuracy across different datasets. They proposed model pruning using TensorFlow's Keras tool, maintaining high accuracy for RML2016.10a/b but showing a drop for RML2018.01a. Guo et al. [17] proposed an SNN-based classification method using $\Sigma\Delta$ spike encoding, achieving up to 64.29% accuracy with 84k- 627k parameters. Li et al. [18] proposed a lightweight multi-feature fusion structure (lightMFFS) CNN architecture for AMC, with an asymmetric convolution structure and attention-based fusion mechanism, achieving 63.44%-65.44% accuracy for RML2016.10a/b with 95k parameters and 0.34 million FLOPs.

### C. Models with 100k to 250k Tunable Parameters

Zheng et al. [19] introduced a Transformer-based automatic modulation recognition (TMRN-GLU) model leveraging CNN and RNN for modulation classification. TMRN-GLU achieved 65.7% average accuracy and 93.7% max accuracy on RML2016.10b with 106k parameters. A smaller version, TMRN-GLU-Small, had 25k parameters with 61.7% accuracy. Ning et al. [20] proposed a transformer-based model, MAMR, leveraging multimodal fusion of I/Q and Fractional Fourier Transform (FRFT) signals to improve classification robustness, achieving an average accuracy of 61.55% and 79.01% on the RML2016.10a and HisarMod2019.1 datasets with 100k model parameters. Its reliance on computationally expensive FRFT transformations remains challenging. Shi et al. [21] proposed a depthwise separable CNN with self-attention, achieving 98.7% maximum classification accuracy on RML2018.01a with 113k parameters. Xue et al.[22] proposed MLResNet, an improved ResNet with an LSTM DL network architecture of parameter size of 115k, achieving a maximum accuracy of 96% at 18dB for the RML2018.01a dataset. Riddhi et al. [23] created a hybrid CNN-GRU model with an attention mechanism. Despite using fewer layers and filters, the architecture still comprises 145k tunable parameters, achieving an accuracy of over 96% at high SNRs and over 75% at lower SNRs (−4 dB to 18 dB) for digital modulation types.

Parmar et al. [24] proposed a dual-stream CNN-BiLSTM model, achieving an average 68.28% accuracy on RML2016.10b for digital modulation types using 145k tunable parameters, demonstrating an effective balance between feature extraction and model size. In [25], Parmar et al. introduced a multilevel classification approach with three DL models (AD-MC, An-MC, and Dig-MC) to classify signals into analog or digital categories and classify specific modulation types. While this approach demonstrated superior performance at higher SNRs, it required 155k parameters and achieved an average accuracy of 63% on the RML2016.10b dataset, balancing complexity with moderate performance improvements. Chang et al. [26] introduced Fast Multi-Loss Learning Deep Neural Network (FastMLDNN), which stacks three group convolutional layers and a transformer encoder to improve feature extraction while minimizing the risk of overfitting. The proposed model reduces computational overhead nearly nine times compared to its baseline, MLDNN [27], achieving 63.24% average accuracy for RML2016.10a with 159k parameters. Luo et al. [28] developed RLITNN for low-SNR recognition, integrating LSTM and transformer-encoder modules, multi-head attention mechanisms, and multiple feature extraction modules for amplitude, phase, spectrum, and power spectral density. The model required 181k parameters and 53.75 million FLOPs, achieving 63.84% (RML2016.10a) and 65.32% (RML2016.10b) average accuracy. Harper et al. [29] focused on differentiable statistical moment aggregation for feature learning, using fixed and learnable moments to refine modulation scheme representations using CNN architecture with squeeze and excitation (SE) blocks. With 200k parameters, it has an average of 63.15% and a maximum of 98.9% accuracy using RML2018.01a with high computational overhead. In [30], Harper et al. integrated CNN architecture with dilated convolutions, statistics pooling, and SE units. It achieved peak accuracy with 202k tunable parameters of 98.9% and an average accuracy of 63.7% on the RML2018.01a dataset. Huynh-The et al. [31] proposed a cost-effective and high-performance CNN model, MCNet. The model with 220k parameters achieves a classification accuracy of over 93% at 20 dB SNR for the RML2018.10a dataset. In [32], Sun and Wang developed a Fusion GRU DL Neural Network (FGDNN) combining GRUs and CNNs to enhance spatiotemporal feature extraction. Their model achieved 90% accuracy at 8 dB SNR with 253k parameters. Nisar et al. [33] utilized ResNet blocks and SE networks to improve AMC's accuracy and efficiency. Their model, with 263,000 parameters and 749 million FLOPs, achieved accuracies of 81% (18dB) using the RML2016.10a dataset.

Table 1 shows the model performance in ascending order by tunable parameter size, including the proposed model presented in this paper for comparison. It can be observed that classification accuracy improves as model complexity increases and vice versa. The proposed model in this paper attempts to minimize the model complexity while maintaining classification accuracy.

## III. Problem Statement, Hypothesis, and Research Question

### A. Problem Statement

The problem is that achieving high classification accuracy in DL models for AMC necessitates increased model complexity. This complexity renders these models unsuitable for deployment on widely distributed edge devices, such as IoT, medical sensors, and smart home systems, which are constrained by limited memory and computational resources. Efforts to reduce DL model complexity often result in compromised classification accuracy, creating a critical challenge for the effective deployment of AMC in edge environments.

### B. Hypothesis

If we can develop a novel DL model for AMC tailored for edge devices, which harmoniously balances high modulation classification accuracy and low model complexity in an overcrowded and noisy wireless environment, then we can effectively tackle the challenge of creating efficient DL AMC models for edge devices.

### C. Research Question

How can advanced lightweight AMC models be designed to optimize classification accuracy and model complexity for resource-constrained edge devices?

## IV. Methodology and Proposed Model

### A. Methodology

Modulating a baseband signal encoded with input data to a high-frequency carrier signal transmitted over the air channel is called modulation. In a typical modern wireless communication system, the transmitted signal is dynamically modulated based on channel conditions and specifications of the system [34]. The primary method of conducting the modulation classification is by deploying a DL classification model on incoming propagation signals at the radio receiver of the wireless system. The objective of AMC can be described mathematically by (1), where $\hat{y}$ represents the predicted modulation type, y represents the actual modulation type, and W represents the learned weights of the DL model.

$$\hat{y} = argmax f(y|X; W) \qquad (1)$$

Incoming propagation signals are represented by $X$ is an input feature that represents in-phase and quadrature-phase (I/Q) baseband signals. In a single-input single-output (SISO) system, the received signal can be expressed as shown in (2):

$$x(t) = s(t) * h(t) + n(t) \qquad (2)$$

Here, $s(t)$ represents the modulated signal, $h(t)$ denotes the channel impulse response and $n(t)$ accounts for white Gaussian noise. The modulation signals take on different forms depending on the type of modulation. To simplify subsequent signal processing and modulation recognition, the received signal is typically represented by in-phase and quadrature (I/Q) components [35].

TABLE 1    MODEL PERFORMANCE BY TUNABLE PARAMETER; DATASETS A: RML2016.10A, B: RML2016.10B, C: RML2018.01A

| Author | Year | Model Name | DL Architecture | Trainable Parameters | Dataset | Avg Accuracy | Max Accuracy |
|---|---|---|---|---|---|---|---|
| Lin et al. [4] | 2024 | | LSM | 7k | A | 36.39% | |
| | | | | | B | 39.74% | |
| | | | | | C | 53.79% | |
| Shaik and Kirthiga[7] | 2021 | | DenseNet | 19k | C | 55% | |
| TianShu et al. [8] | 2022 | IQCLNet | CNN, LSTM | 29k | A | 59.73% | |
| Shen et al. [9] | 2023 | MSSA | CNN | 36k to 218k | C | 55.25% to 60.90% | |
| An et al. [10] | 2023 | TDRNN | CNN, GRU | 41k | A | 63.5% [ -8dB to18dB] | |
| Gao et al. [11] | 2023 | | CNN, LSTM | 43k | A | 66.7% [5 Modulations] | |
| Li et al. [13] | 2024 | CV-TRN | Complex Value Transformer | 44k | A | 63.74% | 93.76% |
| | | | | | C | 64.13% | 98.95% |
| Su et al. [14] | 2022 | SigFormer | Transformer | 44k, 44k, 158k | A | 63.71% | 93.60% |
| | | | | | B | 65.77% | 94.80% |
| | | | | | C | 63.96% | 97.50% |
| Ding et al. [15] | 2023 | | CNN, BiGRU | 69k | C | | 34% (-10dB) [-10dB to 2dB] |
| Zhang et al. [16] | 2021 | PET-CGDNN | CNN, GRU | 71K to 75K | A | 60.44% | |
| | | | | | B | 63.82% | |
| | | | | | C | 63.00% | |
| Guo et al [17] | 2024 | | SNN | 84k - 627k, 83k - 542k | A | 56.69% | |
| | | | | | C | 64.29% | |
| Li et al. [18] | 2023 | LightMFFS | CNN | 95k | A | 63.44% | |
| | | | | | B | 65.44% | |
| Zheng et al.[19] | 2022 | TMRN-GLU (small/large) | Transformer | 25k (Small) 106k (Large) | B | 61.7% (Small) 65.7% (Large) | 93.70% |
| Ning et al. [20] | 2024 | MAMR | Transformer | 110k | B | 61.50% | |
| Shi et al. [21] | 2022 | | CNN, SE block | 113k | C | | 98.70% |
| Xue et al. [22] | 2021 | MLResNet | CNN, LSTM | 115k | C | | 96.6% (18dB) |
| Riddhi et al. [23] | 2024 | | CNN, GRU | 145k | B | 68.23% [Digital Mod. Only] | |
| Parmar et al. [24] | 2023 | | CNN, BiLSTM | 146k | B | 68.23% | |
| Parmar et al. [25] | 2024 | | CNN, LSTM | 155k | B | 63% | |
| Chang et al. [26] | 2023 | FastMLDNN | CNN, Transformer | 159k | A | 63.24% | |
| Luo et al. [28] | 2024 | RLITNN | CNN, LSTM, Transformer | 181k | A | 63.84% | |
| | | | | | B | 65.32% | |
| Harper et al. [29] | 2024 | | CNN, SE Block | 200k | C | 63.15% | |
| Harper et al. [30] | 2023 | | CNN, SE Block | 203k | C | 63.70% | 98.90% |
| Huynh-The et al.[31] | 2020 | MCNet | CNN | 220K | C | | 93% (20dB) |
| Sun & Wang [32] | 2023 | FGDNN | CNN, GRU | 253k | C | | 90%(8dB) |
| Nisar et al. [33] | 2023 | | CNN, SE Block | 253k | A | | 81%(18dB) |
| **Proposed Model** | | **CNN, LSTM, DP-DRSN** | | **27k** | **A** | **61.20%** | **91.23%** |
| | | | | | **B** | **63.78%** | **93.64%** |
| | | | | | **C** | **62.13%** | **97.94%** |

As a result, the I/Q signal can be described as a 2 X N matrix, as shown in (3):

$$X^{IQ} = \begin{bmatrix} X_I\{x[1]\}, X_I\{x[2]\} \dots \dots & X_I\{x[N]\} \\ X_Q\{x[1]\}, X_Q\{x[2]\} \dots \dots & X_Q\{x[N]\} \end{bmatrix} \quad (3)$$

The first row of $X^{IQ}$ corresponds to the in-phase component (I vector), while the second row represents the quadrature component (Q vector). XI and XQ represent the in-phase and quadrature-phase components, respectively, and N is the length of the sample.

In [32], Sun and Wang demonstrated that modulated signal not only have temporal but also has spatial characteristics. They show that the amplitude and phase of the modulated signal improve AMC modulation classification. Thus, Amplitude/phase (A/P) data was generated by converting in-phase/quadrature (I/Q) data from the Cartesian coordinate system to the polar coordinate system shown in (4) and (5). $X^{AP}$ is used as input in the proposed model shown in (6).

$$A[i] = \sqrt{X_I^2[i] + X_Q^2[i]} \quad (4)$$

$$P[i] = \arctan(\frac{X_I[i]}{X_Q[i]}) \quad (5)$$

$$X^{AP} = \begin{bmatrix} A\{x[1]\}, A\{x[2]\} \dots \dots & A\{x[N]\} \\ P\{x[1]\}, P\{x[2]\} \dots \dots & P\{x[N]\} \end{bmatrix} \quad (6)$$

$X^{AP}$ represents a 2 x N vector of amplitude and phase, where a vector represents the amplitude vector, and P represents the phase vector.

### B. Signal Denoising

Wireless signals are inherently susceptible to noise and interference, which can significantly distort the transmitted information and degrade signal quality. This vulnerability necessitates robust signal processing techniques to mitigate noise-induced impairment for reliable demodulation.

In wireless signal denoising, the task is to recover the actual signal coefficient $\theta$ from a noisy observation $x = \theta + \epsilon$, where $\epsilon$ is assumed to be noise. Two estimators used in this context are soft thresholding and garrote thresholding. Soft thresholding is defined by (7)

$$\hat{\theta}_s(x) = \begin{cases} 0, & |x| < \tau \\ (|x| - \tau).Sgn(x), & |x| \geq \tau \end{cases} \quad (7)$$

Which subtracts a fixed threshold $\tau$ from the magnitude of $x$ when $|x| \geq \tau$, resulting in a constant bias of approximately $\tau$ for large true coefficients. In contrast, the garrote thresholding estimator given by (8) can be rewritten as (9).

$$\hat{\theta}_G(x) = \begin{cases} 0, & |x| < \tau \\ \left(x - \frac{\tau^2}{x}\right), & |x| \geq \tau \end{cases} \quad (8)$$

$$\hat{\theta}_G(x) = \begin{cases} 0, & |x| < \tau \\ x\left(1 - \frac{\tau^2}{x^2}\right), & |x| \geq \tau \end{cases} \quad (9)$$

Garrote thresholding operates by scaling x with a factor that depends inversely on the square of x, thereby reducing the bias in proportion to the magnitude of the true signal. A more detailed bias analysis demonstrates that for soft thresholding, when x is close to the true coefficient $\theta$ ($\theta > \tau$). The estimator can be approximated as shown in (10).

$$\hat{\theta}_s(x) \approx \theta - sgn(\theta)\tau \quad (10)$$

This implies a bias of roughly $\tau$. On the other hand, the garrote estimator under the assumption $x \approx \theta$ is approximated by (11) and rewritten as (12).

$$\hat{\theta}_G(x) \approx \theta\left(1 - \frac{\tau^2}{\theta^2}\right) \quad (11)$$

$$\hat{\theta}_G(x) = \theta - \frac{\tau^2}{\theta} \quad (12)$$

Thus, its bias is $\tau^2/_{\theta}$, notably smaller than $\tau$ when $\theta > \tau$. This reduction in bias directly contributes to lower mean-squared error (MSE), as MSE is composed of the squared bias and estimator's variance, as shown in (13).

$$MSE = Bias^2 + Var \quad (13)$$

From the perspective of MSE, soft thresholding's fixed shrinkage introduces a persistent bias that may dominate the error of significant signal components, leading to potential degradation in performance when denoising signals with large coefficients. In contrast, the garrote thresholding's adaptive bias decreases with increasing $\theta$, ensuring that high-magnitude coefficients often present in a wireless communication system that tends to be sparse and bursty are preserved more accurately.

Moreover, as shown in Fig. 2, the smooth transition of the garrote function around the threshold helps to avoid the abrupt discontinuities inherent in soft thresholding, which can otherwise generate artifacts in the denoised signal. Overall, in wireless signal denoising, where preserving a large magnitude, informative coefficient is essential, the garrote thresholding method demonstrates a mathematical advantage over soft thresholding. Its reduced bias for significant components, consequent improvement in overall MSE, and smoother functional behavior justify its preference for applications that demand high fidelity in signal recovery.

Additionally, the DP-DRSN is employed, which enhances denoising effectiveness by leveraging both global average pooling (GAP) and global maximum pooling (GMP) to estimate adaptive thresholds. In contrast to single-path deep residual shrinkage network (SP-DRSN) that rely solely on GAP, DP-DRSN both capture the average contextual information and the prominent high-amplitude signal features that often signify transient noise. The denoising threshold is computed using a learnable convex combination as shown in (14).

$$Threshold = \kappa * (\gamma * \alpha + (1 - \gamma) * \beta) \quad (14)$$

Where, $\alpha$ and $\beta$ are derived from GAP and GMP, respectively, $\gamma \, \epsilon \, [0,1]$ is a learnable weight, and self learning $\kappa$ scales the combined threshold. This formulation allows the model to dynamically adjust denoising sensitivity during training, thereby enhancing robustness in heterogeneous and noisy environments.

### C. Proposed Model

The proposed model has three blocks: feature extraction, denoiser, and classifier. The choices in the model layer parameters are intended to minimize the model's weight while maximizing its accuracy.

The feature extraction block is a hybrid model composed of LSTM and CNN layers that extract the temporal and spatial features of I/Q and A/P signals (refer to Fig. 1). Temporal features are extracted using the LSTM layer with four units while the returning sequence is true. The output of the LSTM layer was reshaped to be concatenated with the CNN block with a filter size of 4. Spatial features were extracted using the asymmetric convolution [18] layers, using a filter size (3x1), extracting horizontal features, and (1x3) extracting vertical features. This decomposition reduces the number of parameters and speeds up computation. The horizontal and vertical features are concatenated along axis 2. In addition, the CNN layer used a dilation factor of (2,2) to capture global features without increasing computational cost. CNN layers included a kernel initializer of he_normal and the kernel regularizer of L2 with a regularization factor of 1e-04 to facilitate model convergence. Finally, temporal and spatial features extracted from the LSTM and CNN layers are concatenated along axis 3.

The theory to denoise the complex and noisy signals is the continuous wavelet transform (CWT). CWT has successfully been used for image compression, denoising, and signal processing [36]. In this context, CWT's ability to denoise the input signal is of prime importance because of its simultaneous time and frequency representation. A time-frequency transform is required in radio signal propagation because the received signal is a non-stationary time series, and a simple Fourier transform is insufficient. CWT is a convolution of the input signal to transform the input data into a time-frequency transform, which facilitates the selection of dynamic thresholds to denoise signals. In a seminal paper, Donoho [37] proposed a thresholding or shrinkage function to denoise noisy signals in the wavelet transform domain. For effective signal denoising in wavelet thresholding, filters must convert useful data into strong features while minimizing noise. Traditional filter design is complex and requires expertise. DL simplifies this by using gradient descent to learn filters automatically. Thus, integrating soft thresholding with DL effectively removes noise and enhances feature discrimination. Zhao et al. [38] modified a residual network [39], a modification of the convolutional neural network (CNN), to alleviate the vanishing gradient issue, called a deep residual shrinkage network (DRSN), to denoise interference signals as input using soft thresholding for fault diagnosis classification of the mechanical transmission system. Within the context of AMC, An et al. [10] used soft thresholding to denoise the I/Q input signal using a hybrid model with CNN and GRU layers.

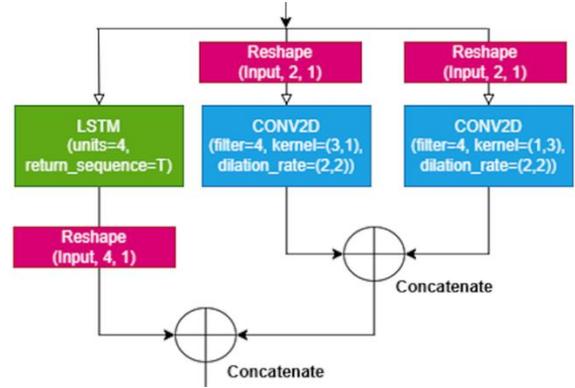

Fig. 1   Feature extraction block

Salimy et al. [40] extended the work in [38] to develop DP-DRSN as a signal-denoising method using a soft thresholding mechanism for fault classification in high-voltage power plant applications. Ruan et al. [41] used DP-DRSN with Soft Thresholding for side-scan sonar image classification. Inspired by their work, this paper uses DP-DRSN for wireless modulated signal classification. We use Garrote Thresholding instead of Soft Thresholding because it is designed to address the limitations of the soft threshold method. This technique employs a non-linear approach for values outside its specified threshold parameter range [42]. Values within the range are set to zero, similar to a soft threshold, while those outside are adjusted non-linearly, as depicted in (15).

$$y_i = \begin{cases} 0, & |x| < \tau \\ x - \dfrac{\tau^2}{(x + 1e - 06)}, & |x| \geq \tau \end{cases} \quad (15)$$

where, x, $y_i$ represents the input and output features and $\tau$ represents the threshold value. Threshold, $\tau$ is a positive parameter; thus, to avoid negative values, a modification is made to add a 1e-06 value to $x$ in the denominator of the Garrote function in the equation above. The process of Garrote Thresholding is shown in Fig. 2. It can be observed that the derivative, referring to (16), of the input is zeroed out for near-zero values.

$$y'_i = \begin{cases} 0, & |x| < \tau \\ 1 + \dfrac{\tau^2}{(x + 1e-06)^2}, & |x| \geq \tau \end{cases} \quad (16)$$

In modulation classification problems, it's crucial to recognize that different modulation types exhibit varying noise levels [43]. Therefore, adjusting the threshold based on the noise level is essential. Prior studies have empirically derived fixed threshold scaling methods for wireless signal-denoising [10] [44]. In contrast, this paper introduces a self-learnable scaling mechanism integrated into the model optimization process, enabling dynamic adjustment of threshold values during denoising. This approach is particularly critical for real-world applications, where devices encounter highly variable noise conditions in increasingly congested dynamic wireless communication environments.

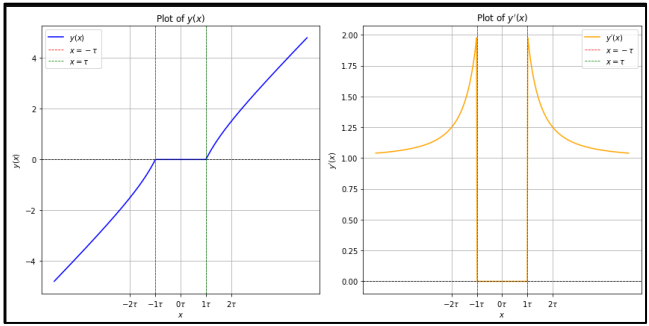

Fig. 2    Garrote Thresholding and its derivative

The structure of DP-DRSN using Garrote Thresholding is shown in Fig. 3. DP-DRSN has a subnetwork, which is an intermediate feature mapping $x \in R^{C \times W \times 1}$ as input. It uses GAP and GMP to the absolute values, extracting the global compressed feature quantity of the current feature map. Subsequently, the 1-D vector is fed into two fully connected (FC) layers with batch normalization and rectified linear unit (ReLU) to derive a scaling parameter [44]. A sigmoid function is applied at the end of the two-layer FC network to ensure the scaling parameter falls within the range of (0, 1), as shown below in (17) and (18).

$$\alpha = \frac{1}{1 + e^{-z_{path1}}} \quad (17)$$

$$\beta = \frac{1}{1 + e^{-z_{path2}}} \quad (18)$$

where, $z_{path1}$ is the output of the two FC layers in the DP-DRSN derived from GAP and $z_{path2}$ is the output of the two FC layers in the DP-DRSN derived from GMP. The proposed model determines the threshold by applying auto-scaling $\tau$ as shown in (19) and (20).

$$\tau = \gamma * \alpha + (1 - \gamma) * \beta, \qquad \gamma = [0,1] \quad (19)$$

$$Threshold = k * \tau \quad (20)$$

Where k, $\gamma$ are self-learning scalar parameters based on the noise level and modulation type. And, $\gamma$ is constrained to the value between 0 and 1. Both values of k and $\gamma$ are optimized through backpropagation during the training process to minimize the loss function. In addition, the kernel initializer of he_normal and the kernel regularizer of L2 with a regularization factor of 1e-04 were used for Conv2D and FC layers to improve training convergence.

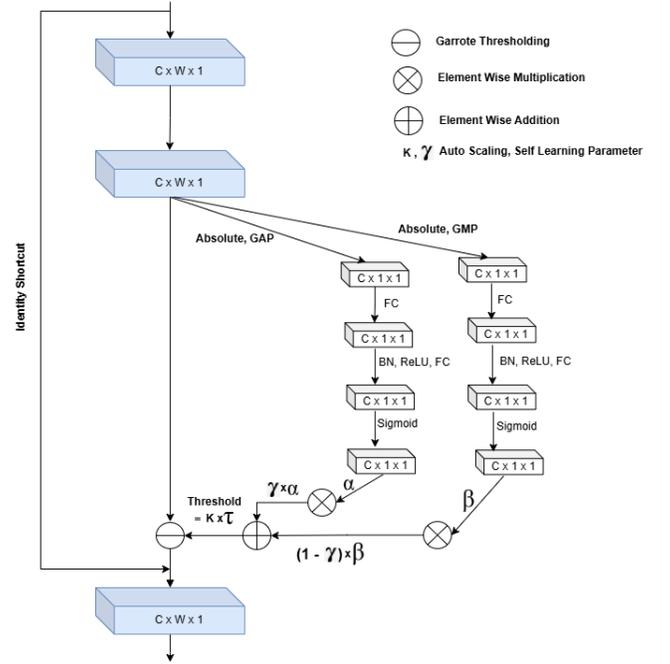

Fig. 3    Dual Path Deep Residual Shrinkage Network (DP-DRSN)

The proposed model's denoiser block comprises four DP-DRSN layers, which facilitate the learning of discriminative features through various nonlinear transformations. Garrote Thresholding functions act as shrinkage mechanisms to remove noise-related information, as shown in Fig. 4.

The Adam optimizer is employed with the categorical cross-entropy (CCE) function as the loss function, shown in (21), for the optimization process. The classifier block module comprises batch normalization, followed by ReLU and GAP operation before applying SoftMax activation as shown in (22).

$$L_{CCE} = \sum_{i}^{N} y \log(\hat{y}) \qquad (21)$$

$$\hat{y} = f_{softmax(x_i)} \qquad (22)$$

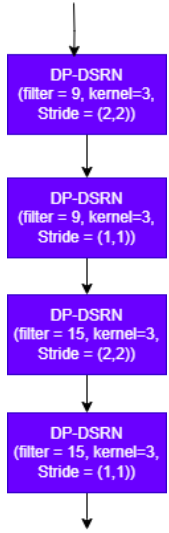

Fig. 4    Denoise block - 4 layers of DP-DRSN

Where $\hat{y}$ represents predicted modulation and $y$ represents the ground truth modulation value with $x_i$ is $i^{th}$ AMC output. Fig. 5 shows the complete proposed architecture of the model. The Algorithm flow is shown in Table 2. The model takes both I/Q and A/P as inputs. The first stage involves extracting spatial and temporal features using CNN and LSTM layers in the feature extraction block for I/Q and A/P inputs. Next, the denoise block minimizes noise features. After that, the denoised I/Q and A/P features are concatenated, and the classifier module identifies modulation classes.

## V. EXPERIMENTS AND RESULT ANALYSIS

### A. Dataset

There are many publicly available datasets that researchers have used to conduct prior AMC research, such as the narrowband radio machine learning (RML) dataset proposed by O'Shea et al., RML2016.10a, RML2016.10b [5], RML2018.01a [6], and HisarMod2019.1 [45]. This paper uses RML2016.10a, RML2016.10b, and RML2018.01a to compare the performance of the proposed model to prior research. Table 3 provides details of datasets with training, validation, and testing configurations.

Additionally, RML2018.01a is used for further testing to evaluate parameter size, FLOPs, energy usage, and classification benchmarking. The RML2018.01a dataset is the prominent benchmark for modulation recognition. Its broad diversity and representative qualities make it ideal for AMC research. The RML2018.01a dataset brings in newer modulation classes, increased signal complexity, and a wider SNR range, keeping pace with advancements in communication technology

TABLE 2    ALGORITHM FLOW - DL MODEL FOR AMC USING DP-DRSN

- *Inputs:*
  - $X_{IQ}, X_{AP} \in \mathbb{R}^{L \times 2}$
  - L : *Signal Length*
  - C: *Number of modulation classes*
  - $\lambda \in [0, 1]$: *Learnable scalar*
  - $\kappa \in \mathbb{R}^{+}$: *Learnable scalar*

- *Feature Extraction per Input: $X \in \{X_{IQ}, X_{AP}\}$*
- *LSTM Path: $H_{LSTM} = Reshape(LSTM_4(X)) \in \mathbb{R}^{L \times 4 \times 1}$*
- *CNN Path:*
  - $X_{reshaped} = Reshape(X) \in \mathbb{R}^{L \times 2 \times 1}$
  - $F_1 = Conv2D_{3 \times 1}^{dil=2}(X_{reshaped}), F_2 = Conv2D_{1 \times 3}^{dil=2}(X_{reshaped})$
  - $H_{CNN} = Concat_{axis=2}(F_1, F_2)$
- *Combined Feature:*
  - $H_{in} = Concat_{axis=3}(H_{CNN}, H_{LSTM})$

- *Dual Path Residual Shrinkage Block with self-learning Garrote Denoising*
- Let $R_0 = H_{in}$
- Step1: Residual Path
  - $R_1 = \varnothing(BN(R_0)) \rightarrow Conv2D$
  - $R_2 = \varnothing(BN(R_1)) \rightarrow Conv2D$
- Step 2: Compute Absolute Value $A = |R_2|$
- Step 3: Compute Global Statistics $\mu = GAP(A) \in \mathbb{R}^C, \nu = GMP(A) \in \mathbb{R}^C$
- Step 4: Calculate the Scaling Coefficient
  - $\alpha = \sigma(\varnothing(BN(Dense(\mu)))), \beta = \sigma(\varnothing(BN(Dense(\nu))))$
  - $\tau = \gamma \cdot \alpha + (1 - \gamma) \cdot \beta$
  - $Threshold = \kappa \cdot \tau(broadcasted\ to\ match\ spatial\ space)$
- Step 5: Apply Garrote Thresholding
  - $M_{small} = \mathbb{I}(A < Threshold), M_{large} = 1 - M_{small}$
  - $R_{small} = M_{small} \cdot 0$
  - $R_{large} = M_{large} \cdot \left(R_2 - \frac{Threshold^2}{R_2 + 1e - 06}\right)$
  - $R_{denoised} = R_{small} + R_{large}$
- Step 6: Skip Connection
  - *If downsample:* $I = AvgPool(R_0)$
  - *If channel($R_0$) $\neq$ channels($R_{denoised}$):* $I = Pad(I)$
  - $R_{out} = R_{denoised} + I$

- *Final Classification:*
- Repeat the above denoising block stack for both IQ and AP paths
- After final residual shrinkage:
  - $F_{IQ} = GAP(\varnothing(BN(R_{final}^{IQ}))), F_{AP} = GAP(\varnothing(BN(R_{final}^{AP})))$
  - $F_{concat} = Concat(F_{IQ}, F_{AP})$
  - $\hat{y} = softmax(Dense_C(F_{concat}))$

and the evolving need for modulation recognition research [46]. RML2018.01a has 24 modulations: 8PSK, BPSK, CPFSK, GFSK, PAM4, AM-DSB, 16QAM, 64QAM, QPSK, WBFM, OOK, 4ASK, BPSK, QPSK, 8PSK, 16QAM, AM-SSB-SC, AM-DSB-SC, FM, and GMSK. A signal dimension of 2 x 1024 and an SNR range of -20dB to 18dB at 2dB increments. There are 2,555,904 samples, 4096 samples for each modulation and SNR, i.e., 24 modulations x 26 SNR levels x 4096 equates to 2,555,904 samples. To simulate real-world scenarios, the signals are distorted with sample rate offset, center frequency offset, symbol rate offset, selective fading, and additive white Gaussian noise (AWGN).

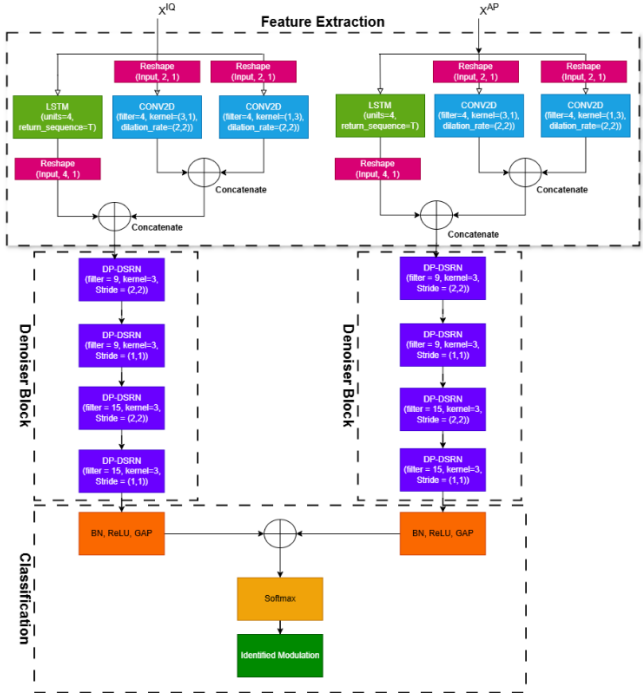

Fig. 5    Proposed model architecture

### B. Experimental Setup

The training process hyperparameters were set to a batch size of 64, 200 epochs, and an initial learning rate of 0.001 using the Adam optimization algorithm. The learning rate is halved if the validation loss remains unchanged after five epochs. Using the EarlyStopping procedure, the training process stops if the validation loss does not decrease in 30 epochs. The experiments were conducted on a Windows 11 machine equipped with an Intel Core i9-2600MHz processor, 64GB RAM, and an NVIDIA GeForce RTX 4080 GPU. Keras and TensorFlow were used to train the model.

TABLE 3    DATASET WITH TRAINING, VALIDATION, AND TESTING CONFIGURATION

| Dataset | Signal Dimension | SNR Range | Train/ Validation/ Test sample size | Modulation classes |
|---|---|---|---|---|
| RML 2016.10a | 2 x 128 | -20dB to 18dB | 132k/44k/ 44k | 11 |
| RML 2016.10b | 2 x 128 | -20dB to 18dB | 720k/240k/ 240k | 10 |
| RML 2018.01a | 2 x 1024 | -20dB to 30dB | 1M/200k/ 200k | 24 |

TABLE 4    PROPOSED MODEL TRAINING PARAMETERS, FLOPs, AND CLASSIFICATION ACCURACY

| Dataset | Trainable Parameter | FLOPs | Avg Acc. | Max. Acc. |
|---|---|---|---|---|
| RML2016.10a | 26,638 | 2.36M | 61.20% | 91.23% |
| RML2016.10b | 26,638 | 2.36M | 63.78% | 93.64% |
| RML2018.01a | 27,072 | 18.71M | 62.13% | 97.94% |

TABLE 5    PROPOSED MODEL MEMORY USAGE, INFERENCE TIME/SAMPLE, AND ENERGY USAGE

| Dataset | Memory Usage (GB) | Inference Time/Sample (ms) | Energy Usage/sample (mJ) |
|---|---|---|---|
| RML2016.10a | 10.47 | 0.14 | 3.31 |
| RML2016.10b | 10.58 | 0.25 | 3.85 |
| RML2018.01a | 10.58 | 1.04 | 31.49 |

### C. Testing and Training Results

The overall model trainable parameters are only 26,638 for the RML2016.10a and RML2016.10b datasets and 27,072 for the RML2018.01a dataset. This lightweight model achieves high performance in terms of average classification accuracy while minimizing the complexity of the model architecture. As shown in Table 4 and Table 5, the proposed model's classification accuracy and model complexity, measured by trainable parameters, FLOPs, inference time per sample (ms), memory usage (GB), and energy usage (mJ) per sample, are provided. Note that by default, TensorFlow allocates all available memory during inference, resulting in consistent memory usage across different models. Overall, the average modulation classification accuracy of 61.20%, 63.78%, and 62.13% was achieved, and maximum classification accuracies of 91.23%, 93.64%, and 97.94% were achieved for RML2016.10a, RML2016.10b, and RML2018.01a, respectively. A training and validation loss plot is shown in Fig. 6. Classification accuracy at each SNR level is shown in Fig. 7.

The proposed model features one of the fewest tunable parameters and fastest inference time, indicating low complexity while achieving high classification accuracy, as shown in Table 6. In addition, a confusion matrix analysis was performed at each SNR level. Fig. 8 shows the confusion matrix at 0dB, and Fig. 9 shows the confusion matrix at 18 dB for RML2016.10a and RML2016.10b, and at 30dB for the RML2018.01a dataset. The other aspect of the DL model complexity evaluation entails the model's inference time, energy usage, and FLOPs. The FLOPs vary based on the size of the input tensor, which, in the case of the AMC model, is based on the signal dimension. The proposed model has 2.36 million and 18.82 million FLOPs for a 2 x 128 and 2 x 1024 signal dimensions, respectively.

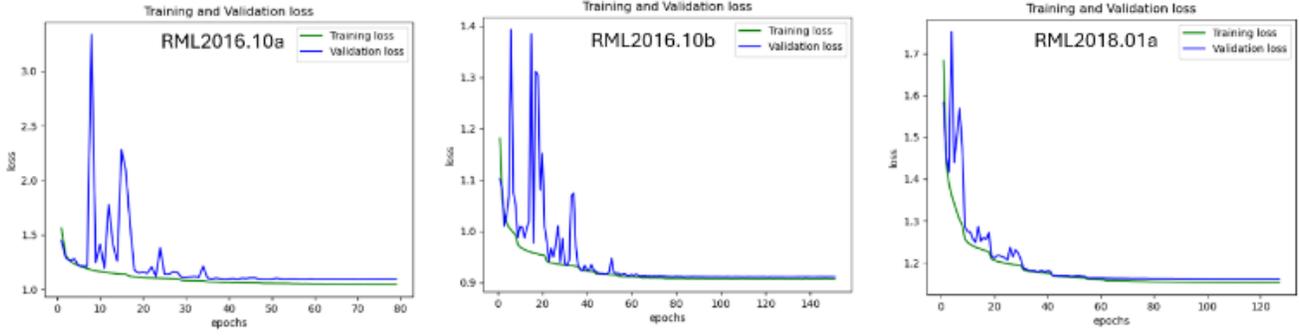

Fig. 6    Proposed model training and validation loss

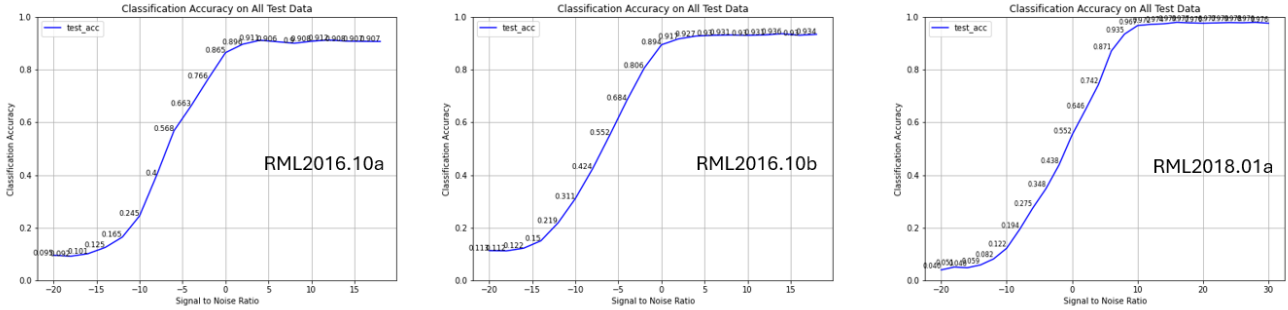

Fig. 7    Proposed model accuracy at each SNR level

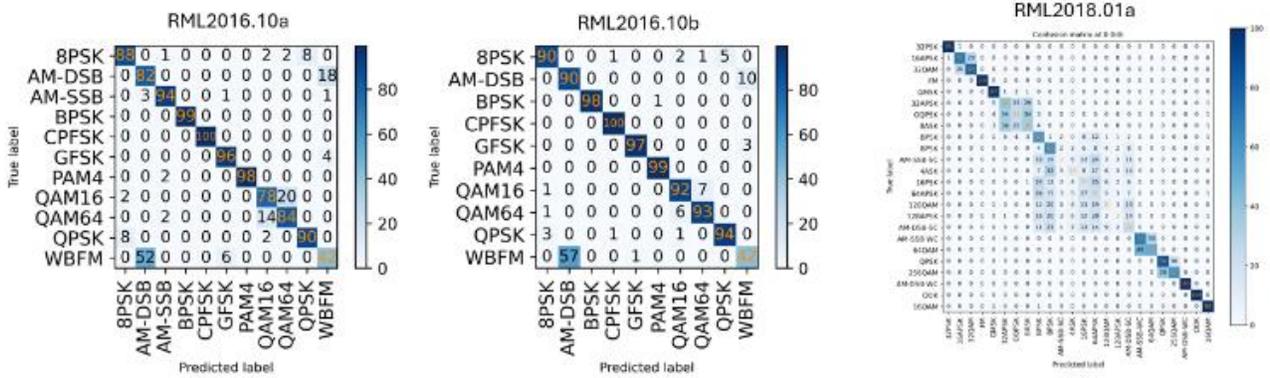

Fig. 8    Confusion Matrix at 0dB

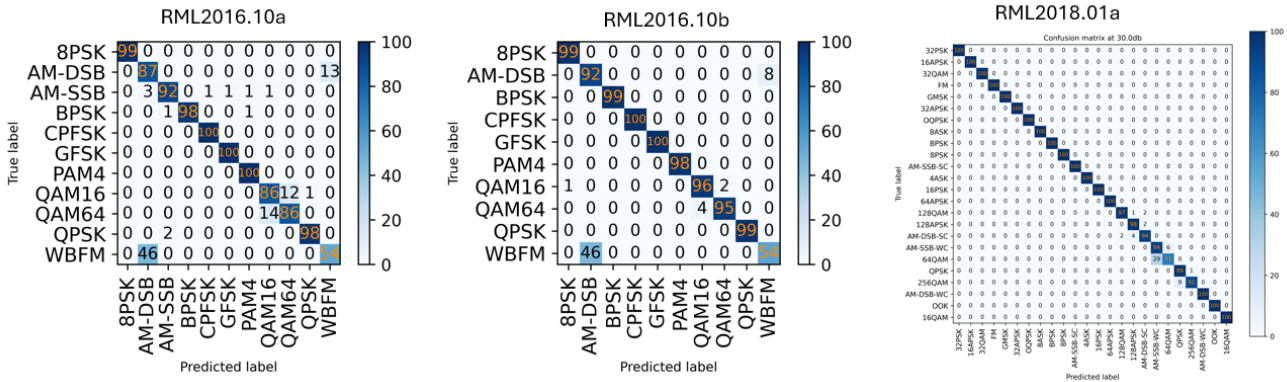

Fig. 9    Confusion Matrix at 18dB for RML2016.10a and RML201610b, and at 30dB for RML2018.01a

In addition, the inference time per sample was recorded as 0.14 ms to 1.04 ms, with energy usage per sample ranging from 3.31 mJ to 31.49 mJ, depending on the signal dimension of the dataset. A state-of-the-art benchmark model comparison is presented in Table 6 for model classification accuracy, FLOPs, inference time, and trainable parameters, focusing on models developed for RML2018.01a reported by [13]. The model proposed in this paper features the fewest parameters and the fastest inference time, while effectively reducing FLOPs, and achieves a balance between classification accuracy and trainable parameter size.

### D. Main Takeaways

The proposed model demonstrates an optimal balance between classification accuracy and model complexity, featuring only 26,638 tunable parameters for RML2016.10a and RML2016.10b and 27,072 for RML2018.01a. With a lightweight architecture, it achieves high classification accuracy, reaching an average accuracy of 62.13%, a maximum accuracy of 97.94%, and an inference time per sample of 1.04 ms for RML2018.01a, while maintaining minimal computational overhead. The model's low inference time, energy consumption, and FLOPs efficiency further highlight its effectiveness. A comparative analysis confirms its superior trade-off between accuracy and resource efficiency, making it a robust solution for modulation classification tasks.

TABLE 6        PERFORMANCE AGAINST BENCHMARK MODELS FOR THE RML2018.01A DATASET

| Model | Avg. Acc. | Max. Acc. | FLOPs (M) | Inference time/Sample (ms) | Param. (K) |
|---|---|---|---|---|---|
| ResNet[6] | 59.29% | 94.12% | 12.18 | 5.40 | 163.22 |
| LSTM[47] | 61.78% | 97.92% | 206.6 | 38.52 | 202.78 |
| MCLDNN [48] | 60.76% | 96.84% | 443.4 | 42.41 | 427.88 |
| PET-CGDNN[16] | 62.52% | 96.14% | 71.92 | 20.63 | 75.34 |
| CDSCNN[49] | 64.13% | 98.01% | 74.11 | 6.80 | 322.62 |
| CV-TRN[13] | 64.13% | 98.85% | 6.42 | 6.81 | 44.79 |
| **Proposed Model** | **62.13%** | **97.94%** | **18.82** | **1.04** | **27.07** |

## VI. DISCUSSION AND FUTURE RESEARCH

### A. Discussion

Additional testing was conducted to validate the performance of the SP-DRSN architecture compared to the DP-DRSN architecture, thereby justifying the increased model complexity using DP-DRSN. The SP-DRSN architecture is shown in Fig. 10. DP-DRSN was replaced by SP-DRSN in the proposed model, using the exact parameters as shown in the Fig. 5 model architecture. The average and maximum classification accuracy declined by 1.7% to 60.53% and 96.36%, respectively. However, the tunable parameters of the proposed model with SP-DRSN were reduced to 24,828, i.e., roughly a 9% reduction, and FLOPs remained the same at 18.81 million. Fig. 11 compares classification accuracy by SNR SP-DRSN vs DP-

DRSN using Garotte Thresholding. This demonstrates that the proposed model with DP-DRSN balances the demand for classification accuracy and model complexity.

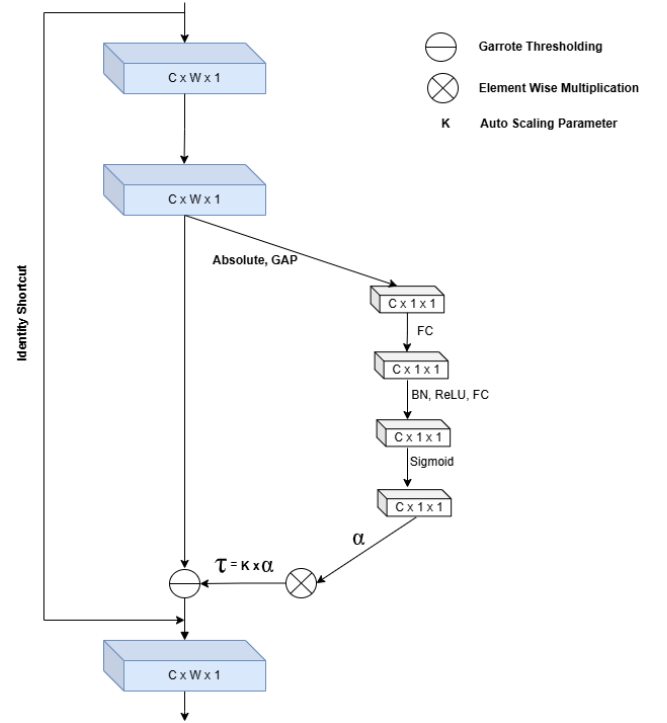

Fig. 10        Single Path DRSN (SP-DRSN)

The proposed model's performance was also evaluated using the soft thresholding denoising technique instead of the Garrote Thresholding. With soft thresholding, the overall average classification accuracy declined from 62.13% to 61.79%, and the maximum classification accuracy declined from 97.94% to 97.81% across 24 modulations of RML2018.01a, indicating that the Garrote Thresholding is slightly better at denoising the radio noise. It is noteworthy that Garrote Thresholding outperforms soft thresholding at SNR levels ranging from -4 dB to 6 dB, with an average improvement of 2.2%, indicating better performance in a noisy environment. A comparison of the performance is shown in Fig. 11 with a zoomed-inset plot.

Equations 23 and 24 illustrate the Soft Thresholding method and its derivative. It can be seen that Soft Thresholding removes near-zero values within the threshold range, similar to Garrote Thresholding; however, those outside the threshold are adjusted linearly (refer to Fig. 12 for Soft Thresholding and its derivative).

$$y_i = \begin{cases} 0, & |x| < \tau \\ (|x| - \tau).Sgn(x), & |x| \geq \tau \end{cases} \quad (23)$$

$$y'_i = \begin{cases} 0, & |x| < \tau \\ 1, & |x| \geq \tau \end{cases} \quad (24)$$

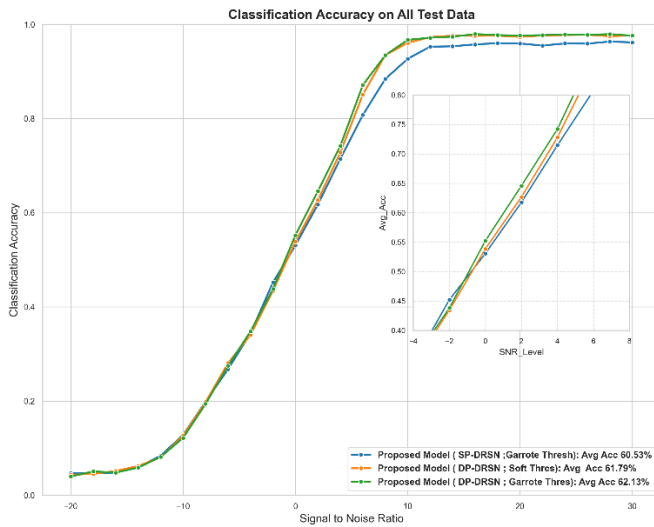

Fig. 11    Proposed model classification accuracy vs SNR for SP-DRSN Garrote Thresholding, DP-DRSN Garrote Thresholding, and DP-DRSN with Soft Thresholding

## B.  Recommendations for Future Research

As the field of AMC continues to evolve, several promising research avenues merit further exploration to enhance their efficacy and applicability. These focus areas include more efficient signal denoising mechanisms to address the ever-evolving overcrowded wireless spectrum, using real-world data, reducing model energy consumption, reducing FLOPs, and model pruning to reduce model complexity. One of the primary areas for future research is the development of advanced denoising mechanisms. The ongoing expansion of wireless communication systems has created a more crowded spectrum, highlighting the importance of effective signal denoising. Future research should aim to develop innovative denoising algorithms that can accurately filter out noise while preserving essential signal characteristics.

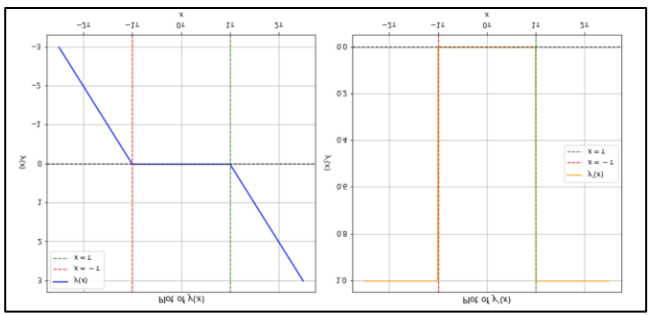

Fig. 12    Soft Thresholding and its derivative

Another critical research direction is using real-world data to train and test AMC models. Real-world data often presents more complex and diverse scenarios than synthesized data, leading to a more robust and generalizable model [50]. Future studies should focus on collecting, processing, and leveraging extensive real-world datasets to enhance the performance of AMC systems in practical applications.

Improving the energy consumption of AMC models is essential for real-time communication applications. Research efforts should optimize the computational efficiency of these models using techniques such as model compression, hardware acceleration, and algorithmic optimizations, which should be explored to achieve faster inference times and lower energy consumption without compromising recognition accuracy. Another critical area of focus is reducing the number of FLOPs required by AMC models. Lowering FLOPs can lead to more efficient models that consume less computational power and energy.

Future research should investigate methods to streamline model architecture, such as employing lightweight neural network designs, such as sparse self-attention, or leveraging quantization techniques. These approaches could result in more resource-efficient AMC systems, making them suitable for deployment on edge devices with limited computational capabilities [51]. Model pruning is a valuable technique for reducing the complexity of AMC models. By selectively removing redundant or less significant parameters from the neural network, model pruning can lead to smaller, more efficient models without compromising performance [16]. Future research should focus on developing advanced pruning algorithms that can effectively balance model size and accuracy. The development of more efficient, accurate, and versatile AMC systems ultimately enhances the overall effectiveness of modern communication networks.

## VII. Conclusion

The proposed approach effectively addresses the challenge of model complexity while maintaining high classification accuracy, making it well-suited for resource-constrained edge devices. The developed model is exceptionally compact, with a size of only 108.83 kB and just 27k tunable parameters. It requires between 2.36 and 18.71 million FLOPs, achieves an inference time of 0.14 to 1.04 ms per sample, and consumes 3.3 to 31.5 mJ of energy per sample. As demonstrated in Table 1, this positions the model among the most efficient in terms of complexity. Despite its reduced complexity, the proposed model achieves high average classification accuracies of 61.20%, 63.78%, and 62.13% on the RML2016.10a, RML2016.10b, and RML2018.10a standard datasets, respectively. This highlights its potential for efficient deployment in real-world edge environments. As a core enabler of intelligent wireless systems, AMC is key to next-generation communication systems, promoting innovation, efficiency, and solutions for a highly connected world.